\pdfoutput=1

\documentclass[11pt]{article}
\usepackage{amsmath}
\usepackage{svg}
\usepackage{listings}
\usepackage{xcolor}
\usepackage{float}
\lstdefinestyle{promptstyle}{
    basicstyle=\ttfamily\small,
    breaklines=true,
    frame=single,
    backgroundcolor=\color{gray!10},
    keywordstyle=\color{blue},
    commentstyle=\color{gray}
}

\usepackage[preprint]{acl}

\usepackage{times}
\usepackage{latexsym}
\usepackage{booktabs}
\usepackage[T1]{fontenc}

\usepackage[utf8]{inputenc}

\usepackage{microtype}

\usepackage{inconsolata}

\usepackage{graphicx}

\title{DebateBench: A Challenging Long Context Reasoning Benchmark For Large Language Models}

\author{{\bf Utkarsh Tiwari*} \quad {\bf Aryan Seth*} \quad {\bf Adi Mukherjee} \\ 
        {\bf Kaavya Mer} \quad {\bf Kavish} \quad {\bf Dhruv Kumar\textsuperscript{\textdagger}} \\
        Birla Institute of Technology and Science, Pilani \\
        \texttt{f20212221, f20220052}@pilani.bits-pilani.ac.in}

\begin{document}
\maketitle
\begin{abstract}
We introduce DebateBench, a novel dataset consisting of an extensive collection of transcripts and metadata from some of the world’s most prestigious competitive debates. The dataset consists of British Parliamentary debates from prestigious debating tournaments on diverse topics, annotated with detailed speech-level scores and house rankings sourced from official adjudication data. We curate 256 speeches across 32 debates with each debate being over 1 hour long with each input being an average of 32,000 tokens. Designed to capture long-context, large-scale reasoning tasks, DebateBench provides a benchmark for evaluating modern large language models (LLMs) on their ability to engage in argumentation, deliberation, and alignment with human experts. To do well on DebateBench, the LLMs must perform in-context learning to understand the rules and evaluation criteria of the debates, then analyze 8 seven minute long speeches and reason about the arguments presented by all speakers to give the final results. Our preliminary evaluation using GPT o1, GPT-4o, and Claude haiku, shows that LLMs struggle to perform well on DebateBench, highlighting the need to develop more sophisticated techniques for improving their performance.
\end{abstract}

\section{Introduction}

\def\thefootnote{*}\footnotetext{Equal contribution}\def\thefootnote{\arabic{footnote}}
\def\thefootnote{\textdagger}\footnotetext{Senior author}\def\thefootnote{\arabic{footnote}}

The reasoning capabilities of Large Language Models (LLMs) have been extensively evaluated across a variety of domains, including STEM problem-solving \cite{cobbe2021trainingverifierssolvemath, clark2018thinksolvedquestionanswering, arora2023llmsadvancedenoughchallenging, hendrycks2021measuringmathematicalproblemsolving, lu2022learn, bubeck2023sparksartificialgeneralintelligence}, language understanding \cite{hendrycks2021measuringmassivemultitasklanguage}, and code generation \cite{chen2021evaluatinglargelanguagemodels, austin2021programsynthesislargelanguage}. However, there remains a notable gap in the availability of sufficiently diverse and challenging natural language datasets that rigorously benchmark reasoning over long-contexts. Additionally, existing benchmarks for long-context reasoning suffer from two main problems: (1) lack of "argument-only" debates, and (2) non-comprehensive scoring metrics. 
\begin{figure}[t]
    \centering
    \includegraphics[width=\linewidth]{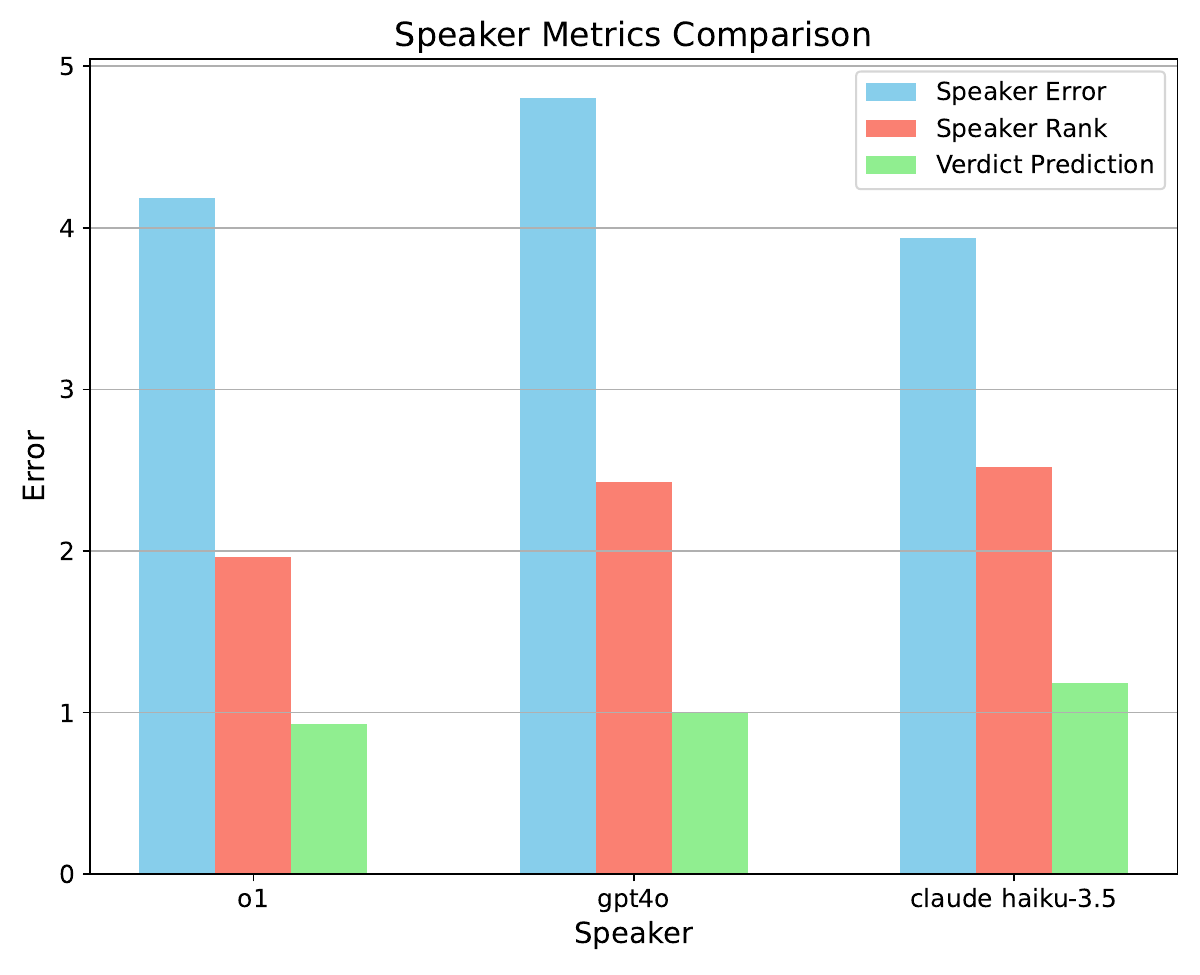}
    \caption{Performance of models on DebateBench, the y-axis represents the mean absolute error (MAE) of the three tasks. More details in \ref{section:eval}}
    \label{fig:results}
\end{figure}
Argument-only debates differ from factual debates in that the judging panel knows no facts beforehand and needs to be convinced of their correctness. Existing metrics in these datasets are also win/loss based, rather than provided by expert annotators.  A comprehensive overview of existing benchmarks is provided in Section \ref{related}.

In this paper, we propose DebateBench, a novel dataset consisting of an extensive collection of transcripts and metadata of British Parliamentary debates from prestigious debating tournaments on diverse topics. It is an annotated dataset containing detailed speech-level scores and house rankings sourced from official adjudication data. We curate 256 speeches across 32 debates with each debate being over 1 hour long with each input being an average of 32,000 tokens. DebateBench serves as a challenging benchmark for evaluating modern large language models (LLMs) on their ability to engage in argumentation, deliberation, and alignment with human experts.


DebateBench includes three primary evaluation tasks: (1) \textbf{Speech Scoring}, where models predict human-assigned scores for individual speeches, (2) \textbf{Speech Ranking}, where the model predicts the speaker rankings, and (3) \textbf{House Ordering}, where models rank debating teams by adjudication outcomes.

This paper makes three primary contributions:
\begin{enumerate}

\item \textbf{New Dataset}: We introduce, DebateBench, a long-context reasoning benchmark comprising 32 debates conducted in the British Parliamentary format, approximately having 100,000 words per debate. 

    \item \textbf{Novel Task:} We introduce complex multi-turn debates for reasoning and structured format argumentation over long contexts, scored against a human ground truth.

    \item \textbf{Evaluation}: Top-of-the-line LLMs struggle on DebateBench, demonstrating their inability to handle dense long-context tasks that require structured argumentation. Additional details are discussed in Section 4.
\end{enumerate}
\section{Related Works} \label{related}
Reasoning in LLMs has been studied under multiple contexts such as logical reasoning, mathematical reasoning, theorem proving etc. In real life contexts, this extends to fields at the intersection of reasoning, decision-making, and communication, such as law, politics, and education. In this section we summarize the literature on natural language reasoning which is the closest related to our task.

\subsection{Natural Langauge Reasoning}
Natural Language datasets like \textbf{HellaSwag} \citep{zellers2019hellaswagmachinereallyfinish},  \textbf{ARC} \citep{clark2018thinksolvedquestionanswering}, and \textbf{MMLU} \cite{hendrycks2021measuringmassivemultitasklanguage} test LLMs ability to understand natural language and ground their answers in reality. HellaSwag tests models' ability to complete sentence by choosing the most likely option from 4 sentences provided, while ARC and MMLU test models on either domain dependent information or common sense reasoning by asking questions on biology, law, economics, etc. The \textbf{TruthfulQA} \citep{lin2022truthfulqameasuringmodelsmimic} benchmark is especially designed to test models' grounding in reality by evaluating them on questions who's answer are prone to be misinformation or conspiratorial. 

Other benchmarks like \textbf{SuperGLUE} \citep{wang2020supergluestickierbenchmarkgeneralpurpose} and \textbf{WinoGrande} \citep{sakaguchi2019winograndeadversarialwinogradschema} test models' comprehension of natural languages by testing them on confusing or ambiguous passages.

More closely aligned to our work are argument evaluation benchmarks like \textbf{VivesDebate-Speech} \citep{ruizdolz2024vivesdebatespeechcorpusspokenargumentation}, which is a dataset of 29 debates from the 2019 university debate tournament organized by the “Xarxa Vives d’universitats”. However, the debates in this benchmark are not originally English, and have been machine translated from Catalan. The credibility of machine translations in preserving complicated arguments is low. Moreover, DebateBench includes debates from renowned competitions hence the judges' scores are more credible and the debates are of a higher quality. Other "debating" datasets like \textbf{USElecDeb60To16} \citep{haddadan-etal-2019-yes}, and \textbf{ETHIC} \citep{lee2024ethicevaluatinglargelanguage} benchmark deal with political debates between U.S. Presidential candidates and in the British Parliament respectively. These debates are significantly different from competitive debates since the main focus is on rhetoric and not logical argumentation. These debates also don't have a quantifiable metric of evaluation. The \textbf{DebateSum} \citep{roush-balaji-2020-debatesum} deals with \textit{Policy Debates} wherein the topics are released as much a year ago and the competition focuses on the presentation of evidence and data instead of principled arguments. 





\subsection{Long-Context Modeling Techniques}
Recent advancements in LLMs have integrated sophisticated long-context modeling techniques. For instance, LLaMa 2 employs Rotary Position Embedding (RoPE) \cite{Touvron2023}, while Vicuna 1.5 \cite{Zheng2023} fine-tunes LLaMa 2 to extend context lengths to 16,000 tokens. Similarly, ChatGLM2-32k achieves a 32,000-token context window, demonstrating the scalability of these methods. State-of-the-art models like GPT-4-Turbo (128,000 tokens) and Claude-3.5-Sonnet (200,000 tokens) further push the boundaries of context length, enabling the processing of extensive information. Despite these advancements, there is a notable scarcity of human-aligned benchmarks designed to evaluate performance at such scale. 
\section{The DebateBench Dataset}

\begin{figure*}[t]
    \centering
    \includegraphics[width=0.65\textwidth]{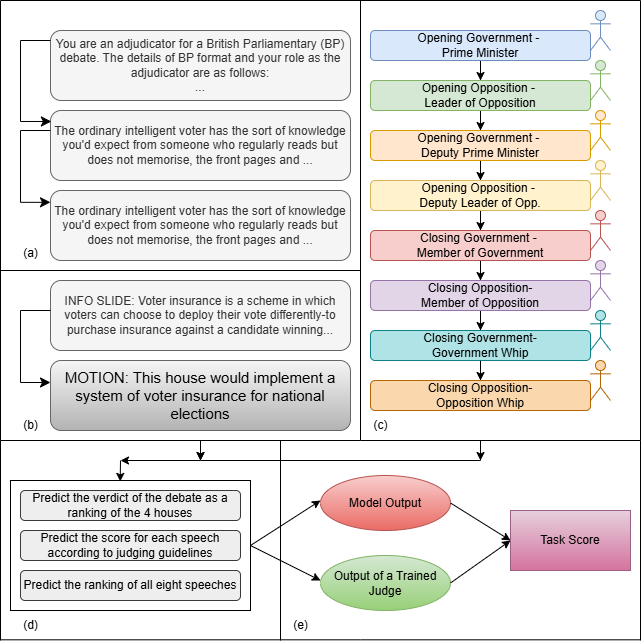}
    \caption{The system prompt explaining the format of the debate as well as the metrics of judgment (a) along with the information slide (if present) and the motion (b) and the transcript of the debate (c) which contains 8 speeches by 4 teams (or houses) is passed to the model. The model is tested on 3 tasks (d) and the output is compared to the results given by trained judges to compute the task scores for the model (e).}
    \label{fig:example}
\end{figure*}

The dataset comprises 32 debates, each consisting of 8 speeches approximately 7 minutes long, curated from prestigious tournaments such as the Doxbridge Worlds Schools Debating Championships, LSE Opens, and past editions of the World Universities Debating Championship (WUDC). These tournaments are recognized for their high-quality debates, ensuring a robust benchmark. A full list of debates is provided in the Appendix.

Debates follow the British Parliamentary format, featuring four teams (or houses): Opening Government, Opening Opposition, Closing Government, and Closing Opposition. Each team competes in a structured round, with detailed format specifications provided in Figure \ref{fig:example}.

The debate topic, referred to as the "motion," is framed as a proposal for "the house." For example, a motion from the Cambridge IV 2020 debate states: "This house believes that protest movements should actively integrate religious figures and institutions in opposition to authoritarian regimes." Government teams support the motion, while opposition teams oppose it. Teams are allotted 15 minutes of preparation time without internet access, discouraging reliance on statistics or specialized knowledge. Instead, speakers are expected to construct arguments based on principles and plausible scenarios, logically extending them. Judges, guided by the WUDC judging manual\footnote{\url{https://shorturl.at/QnjKe}}, are instructed to evaluate debates as "ordinary intelligent voters" or "informed global citizens." They discount appeals to highly specialized concepts unless clearly explained, ensuring arguments remain accessible. While complex claims are permissible, they must be articulated in jargon-free, understandable terms. During the evaluation, LLMs are provided with the judging manual and instructed to adjudicate accordingly (see Appendix for the full system prompt).

The judging manual also outlines heuristics for speech objectives. For instance, the first two speeches are expected to contextualize the debate and clarify ambiguities, while the final two speeches must identify major clashes and justify their team's success. The system prompt, adapted from the WUDC manual, exceeds 18,000 tokens and includes details on debate format, speaker roles, and adjudication heuristics. Additionally, most motions include an "Info Slide" explaining key terms, as illustrated in Figure \ref{fig:example}. Consequently, performing well on DebateBench requires models to excel at in-context learning \cite{dong2024surveyincontextlearning}, further underscoring the benchmark's complexity.

\subsection{Dataset Collection}
    We first curate debating videos from YouTube. These videos are recorded with prior consent of all speakers and publicly shared. We remove the videos which contain less than 8 speeches, and those in which large sections are unintelligible due to bad audio quality. We then use GPT-Whisper \citep{verma2024whisper} to generate transcripts. However, these transcripts are of low quality and need to be processed before being used. We then use GPT-4o mini to correct any grammatical errors and spelling mistakes. The timestamps are preserved in this step.
    It was observed, that this step was able to correctly infer punctuation and correct grammar and spelling.

    Finally, all the transcripts are manually verified to ensure a high quality dataset. Speaker tags and other tokens are added to differentiate speakers in the debate.

    \begin{table}[h]
    \centering
    \begin{tabular}{l r}
        \toprule
        Statistic & Value \\
        \midrule
        Total number of speeches & 256 \\
        Total number of words & 884,395 \\
        Total number of tokens & 1,326,592 \\
        Average number of words per speech & 11,904 \\
        Total hours of content & $\sim$36 \\
        Average speaker score & 80.25 \\
        Speaker score standard deviation & 2.69 \\
        \bottomrule
    \end{tabular}
    \caption{Statistics of the dataset. The tokens are calculated assuming the standard 1 word $\approx$ 1.5 tokens.}
    \label{tab:dataset_stats}
\end{table}

\subsection{Evaluation Metrics}
\label{section:eval}
We propose three tasks for evaluating the performance of large language models (LLMs) on DebateBench. For each task, the models are provided with the judging manual along with the debate transcript:

\begin{itemize}
    \item \textbf{Verdict Prediction}: The models are tasked with predicting the ranking of the four houses. The predicted ranking is compared to the ground truth ranking generated by the trained judges by computing the absolute difference at each position. Let the predicted ranking be denoted as \( r_{\text{pred}} = (r_1, r_2, r_3, r_4) \) and the ground truth ranking as \( r_{\text{gt}} = (g_1, g_2, g_3, g_4) \), where \( r_i \) and \( g_i \) are the positions of the houses in the predicted and ground truth rankings, respectively. The verdict prediction score, \( \Delta \), is defined as:
    \[
    \Delta = \sum_{i=1}^{4} |r_i - g_i|
    \]
    If the model predicts the ranking correctly, then \( \Delta = 0 \).

    \item \textbf{Speaker Scores}: Each of the 8 speeches is assigned a score between 50 and 100 by the judges. The guidelines for assigning these scores are provided in the judging manual and passed as a system prompt. This task is formulated as a regression problem, where the model predicts a score \( s_i \in [50, 100] \) for each speech \( i \). The model’s prediction \( \hat{s}_i \) is compared to the true score \( s_i \) by taking absolute difference.
    Additionally, we introduce tolerance \( \epsilon \), allowing the predicted score to fall within an acceptable range of the true score. We ablate on multiple tolerance values, ranging from 2 to 9.

    \item \textbf{Speaker Ranks}: Given the challenging nature of the speaker scoring task, we introduce a simpler extension by asking the model to rank the 8 speeches. Let \( r_{\text{pred}} = (r_1, r_2, \dots, r_8) \) represent the predicted ranks, where \( r_i \) denotes the rank of speech \( i \), and \( r_{\text{gt}} = (g_1, g_2, \dots, g_8) \) denotes the true ranks. We evaluate model performance using the absolute difference metric, defined as:
    \[
    \Delta_{\text{rank}} = \sum_{i=1}^{8} |r_i - g_i|
    \]
    A lower \( \Delta_{\text{rank}} \) indicates a better match with the ground truth ranking. Models that perform well on the speaker scoring task should also perform well on this ranking task. However, this task offers greater differentiation in cases where models perform poorly on speaker scores.
\end{itemize}

\section{Experiments}
\subsection{Experimental Setup}

\paragraph{Settings}We utilize the official WUDC judging manual as context and pass the debate transcripts for evaluation. We apply the corresponding chat and system prompts for each LLM and keep the temperature set to 0.0 while retaining all other sampling parameters as the standard configurations.
\paragraph{Models} We evaluate 3 LLMs, namely OpenAI's GPT4o\cite{openai2024gpt4technicalreport} and o1\cite{openai}, and Anthropic's Claude Haiku 3.5.
\subsection{Main Results}

From \ref{fig:results} we can see that o1 performs the best across the three models for ranking tasks but is still considerably unreliable, with high errors for the verdict and speaker ranking prediction. gpt4o and Haiku 3.5 have comparable results regarding verdict prediction, with all three having an average error of 1.
However, they have varying results on speaker scores, with Haiku performing the best and gpt4o performing the worst. This demonstrates that current LLMs are unable to handle contexts of this size when detailed reasoning is involved.

\subsection{Speaker Score Accuracy}
To evaluate how close LLMs are in predicting speaker scores, which is well demarcated in the WUDC judging manual, we evaluate their performance by checking whether they were a certain tolerance away from the ground truth.
\begin{figure}
    \centering
    \includegraphics[width=\linewidth]{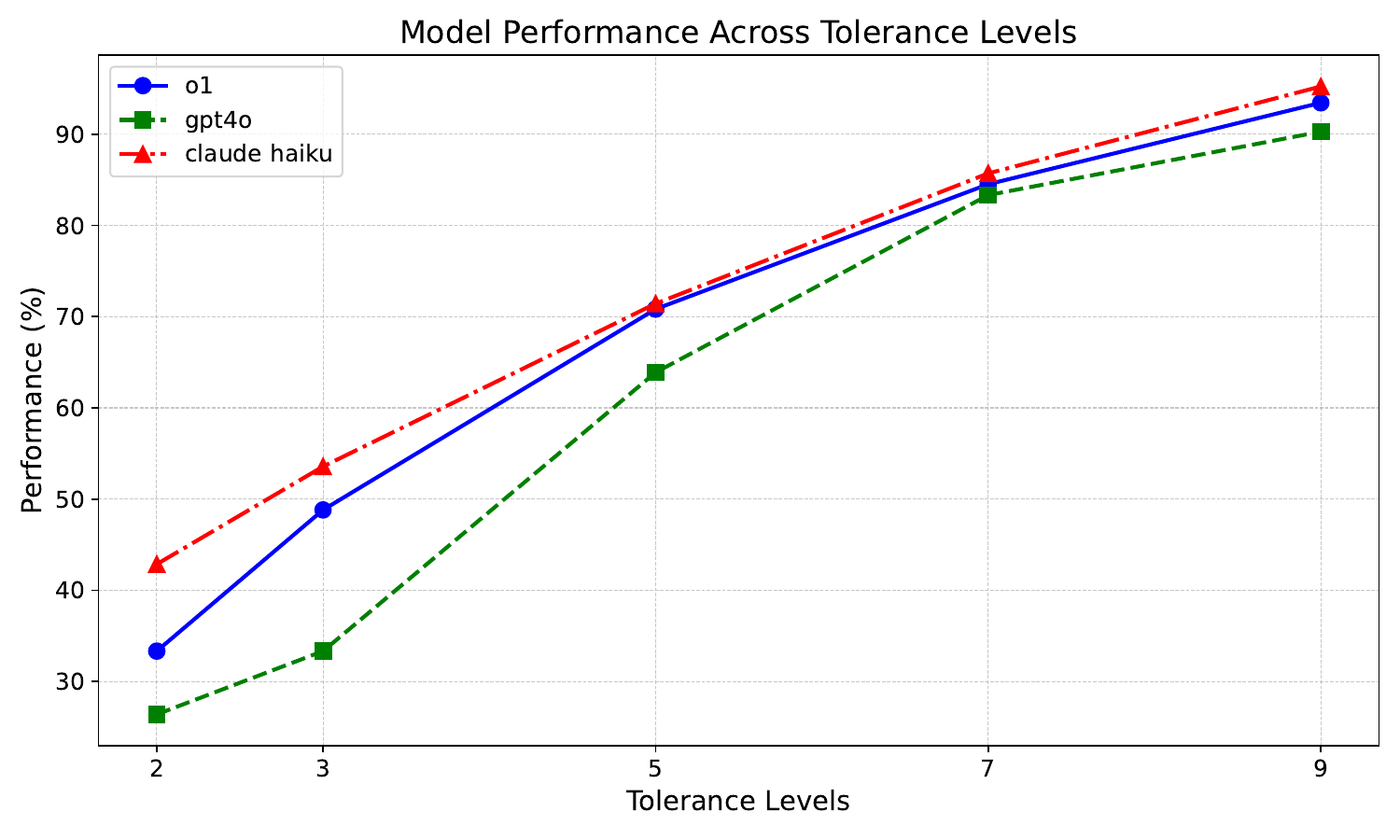}
    \caption{Model accuracy for speaker score prediction at varying delta windows from ground truth}
    \label{fig:tolerance}
\end{figure}
From Figure \ref{fig:tolerance} we see that Claude Haiku performs well, however even at a tolerance of 5, models are about 70\% accurate for score prediction. While this may seem good, it is worth noting that the speaker scores have a standard deviation of 2.69 (as shown in Table \ref{tab:dataset_stats}), which implies that a tolerance of 5 is considerably high.

\section{Future Work and Conclusion}

In this work, we introduce DebateBench, a novel dataset comprising high-quality transcripts and metadata from prestigious competitive debates. This dataset is designed to evaluate language models on their ability to perform in-context learning and logical reasoning over long-form, natural language discourse.

\textbf{Evaluation:} We also present a preliminary evaluation of o1, GPT4o, and Claude Haiku 3.5  on three newly formulated tasks. Our findings indicate that current models struggle to achieve high accuracy on DebateBench, highlighting the challenges posed by its complex reasoning and extensive context requirements.

The limited class of models for evaluation was due to limitations of cost, and future works plan to include a broader range of models featuring varying context windows, different types of preference alignment, and models fine-tuned on this task to evaluate LLMs in a much deeper mannDataset er.

\textbf{Dataset Extension:} A natural extension of this dataset involves incorporating argument annotations into the debate transcripts. Such annotations would enable additional benchmarking tasks, including question-answering, and could serve as a valuable resource for assessing human alignment, as they would be produced by experienced debate judges.

DebateBench is currently formulated to include only those debates for which the metadata defining all three tasks was available. This constraint can be mitigated by expanding the dataset to include debates lacking certain metadata, such as speaker scores. While these debates may not support all evaluation tasks, they will be valuable for verdict prediction and increase the scope of evaluation.

\textbf{Future Work:} Given that debate motions frequently address contentious topics, DebateBench also provides a valuable resource for analyzing bias in language models by examining how they weigh opposing arguments on controversial issues.

Overall, DebateBench presents a challenging benchmark for language model evaluation while simultaneously facilitating future advancements in areas such as human alignment and bias mitigation.


\bibliography{acl_latex}

\newpage

\appendix

\section{System Prompt}

Below is the system prompt for the verdict prediction task:

\begin{figure}[h]
    \centering
    \begin{lstlisting}[style=promptstyle]
You are an experienced debate judge. You have the following judging manual context:
    {judging_manual}

Below is a single text containing all 8 speeches for this round:
    {all_speeches}

Use the judging manual to assign speaker scores in a fair and consistent manner.
Output the rankings in the following format:
First: <house>
Second: <house>
Third:<house>
Fourth: <house>

For example, a sample output could be:
First: OG
Second: CG
Third: OO
Fourth: CO

No additional explanation should be provided.
\end{lstlisting}
    \caption{The judging manual is adapted from the WUDC judging manual and contains 15,361 words. The entire prompt, including the judging manual, can be found in the code repository.}
    \label{fig:enter-label}
\end{figure}

\begin{figure}[h]
    \centering
    \begin{lstlisting}[style=promptstyle]
Astana Open 2023 Round 3
Astana Open Round 4
Astana Open Round 5
Belgrade WUDC 2022 Round 6
Belgrade WUDC 2022 Round 7
Belgrade WUDC 2022 Round 8
Belgrade WUDC 2022 Round 9 Room 2
Cambridge IV 2020 Round 1
Cambridge IV 2020 Round 2
Doxbridge 3 Round 4
Doxbridge 4 Round 1
Doxbridge 4 Round 2
Doxbridge Worlds 2021 West Round 4
LSE Open 2023 Round 2 Room 2
LSE Open 2023 Round 3 Room 1
LSE Open 2023 Round 3 
LSE Open 2024 Round 1A
LSE Open 2024 Round 3 A
LSE Open 2024 Round 3 B
Pakistan Pre ABP 2024 - Round 5
Panama WUDC 2025 Round 1
Panama WUDC 2025 Round 5
Panama WUDC 2025 Round 7
The Natolin European Round Robin Debating Championships Round 1
The Natolin European Round Robin Debating Championships Round 2
The Natolin European Round Robin Debating Championships Round 3
The Natolin European Round Robin Debating Championships Round 4
The Natolin European Round Robin Debating Championships Round 5
Doxbridge 3 Round 3
Doxbridge Pre WUDC 2022 Round 5
Doxbridge Worlds 2021 West Round 5
LSE Open 2023 Round 5
\end{lstlisting}
    \caption{List of debate rounds included in the dataset.}
    \label{fig:dataset_files}
\end{figure}

\label{sec:appendix}

\end{document}